# A ROBUST PARSING ALGORITHM FOR LINK GRAMMARS

Dennis Grinberg      John Lafferty      Daniel Sleator

August 1995

CMU-CS-95-125

School of Computer Science
Carnegie Mellon University
Pittsburgh, PA 15213

Correspondence regarding this paper should be sent to Dennis Grinberg at the above address, or by e-mail to dennis@cs.cmu.edu.

Research supported in part by NSF and ARPA under grant IRI-9314969 and the ATR Telecommunications Research Laboratories.

The views and conclusions contained in this document are those of the authors and should not be interpreted as representing the official policies, either expressed or implied, of the NSF or the U.S. government.

**Abstract**

In this paper we present a robust parsing algorithm based on the link grammar formalism for parsing natural languages. Our algorithm is a natural extension of the original dynamic programming recognition algorithm which recursively counts the number of linkages between two words in the input sentence. The modified algorithm uses the notion of a *null link* in order to allow a connection between any pair of adjacent words, regardless of their dictionary definitions. The algorithm proceeds by making three dynamic programming passes. In the first pass, the input is parsed using the original algorithm which enforces the constraints on links to ensure grammaticality. In the second pass, the total *cost* of each substring of words is computed, where cost is determined by the number of null links necessary to parse the substring. The final pass counts the total number of parses with minimal cost. All of the original pruning techniques have natural counterparts in the robust algorithm. When used together with memoization, these techniques enable the algorithm to run efficiently with cubic worst-case complexity.

We have implemented these ideas and tested them by parsing the Switchboard corpus of conversational English. This corpus is comprised of approximately three million words of text, corresponding to more than 150 hours of transcribed speech collected from telephone conversations restricted to 70 different topics. Although only a small fraction of the sentences in this corpus are "grammatical" by standard criteria, the robust link grammar parser is able to extract relevant structure for a large portion of the sentences. We present the results of our experiments using this system, including the analyses of selected and random sentences from the corpus.

We placed a version of the robust parser on the Word Wide Web for experimentation. It can be reached at URL http://www.cs.cmu.edu/afs/cs.cmu.edu/project/link/www/robust.html. In this version there are some limitations such as the maximum length of a sentence in words and the maximum amount of memory the parser can use.

# 1 Introduction

In this paper we present a robust parsing algorithm for the link grammar formalism introduced in [5]. Using a simple extension of the original formalism we develop efficient parsing and pruning algorithms for extracting structure from unrestricted natural language.

Our approach to robust parsing is purely algorithmic; no modification to the underlying grammar is necessary. We begin by making a generalized definition of what is allowed as a parse. We then assign a non-negative cost to each *generalized parse* in such a way that the cost of a parse which is grammatical with respect to the underlying grammar is zero. The goal of the robust parsing algorithm is then to enumerate all parses of a given sentence that have minimal cost. While this approach to robust parsing is certainly not new, it has a particularly simple and effective realization for link grammar that takes advantage of the formalism's unique properties. In particular, all of the pruning techniques that make link grammar parsing efficient for large natural language grammars either remain unchanged or have natural extensions in the robust parsing algorithm.

The robust algorithm uses the notion of a *null link* to allow a connection between any pair of adjacent words, regardless of their dictionary definitions. The algorithm proceeds by making three dynamic programming passes through the sentence. In the first pass, the input is parsed using the original algorithm which enforces the constraints on links to ensure grammaticality. In the second pass, the total cost of each substring of words is computed, where cost is determined by the number of null links necessary to parse the substring. The final pass counts the total number of parses with minimal cost. Memoization together with pruning techniques enable the algorithm to run efficiently, with theoretic time complexity of $O(n^3)$ for an input of $n$ words.

We have implemented these ideas and tested them by parsing the *Switchboard* corpus of conversational English [1]. This corpus is comprised of approximately three million words of text, corresponding to more than 150 hours of transcribed speech collected from telephone conversations restricted to 70 different topics. Although only a small fraction of the sentences in this corpus are "grammatical" by standard criteria, the robust link grammar parser is able to extract relevant structure for a large portion of the sentences. We present here the results of our experiments using this system, including the analyses of selected and random sentences from the corpus, and the statistics of a typical parsing session.

There is a wide body of literature related to robust parsing. This can be seen immediately by the 200(!) references given by S. Abney in a tutorial titled "Partial Parsing" at the 4th Conference on Applied Natural Language Processing in 1994. While some robust parsing methods depend on specific domains and use semantics as their guide, in this paper we investigate a purely syntactic technique. The advantages of using syntactic techniques include the ability to use the robust parser in varied domains and applications, the ability to use existing grammars with little or no change, and the ability to base techniques on parsing technologies whose characteristics are well understood.

There has been work by a number of researchers on *least-errors recognition* for context-free grammars. In 1974, Lyon [4] proposed a dynamic programming algorithm for finding the least number of mutations, insertions, and deletions of terminal symbols necessary to parse a sentence. Recently, Lee *et al.* [3] extended this work by allowing errors in non-terminals. In [2] Lavie and Tomita describe a modification of the Generalized LR Parser. Their GLR* algorithm is designed to determine the maximal subsets of the input that are parsable by skipping words. While not guaranteed to have cubic running time, a beam search is used to prune parsing options that are unlikely to produce a maximal parse. With pruning, the system is no longer guaranteed to produce the best solution, but the authors report that the beam search works well in practice.

In the following section we briefly review the relevant concepts from link grammar that are necessary for presenting the robust parsing algorithm. In Section 3 we give two definitions of cost. The first is given in terms of the edit distance of a generalized parse and it is analogous to least-errors recognition. The second definition introduces the concept of a null link. Using this definition our approach bears similarity to the work in [2]. While the edit distance is perhaps a more general and intuitive concept, it does not lead us to efficient parsing algorithms. In Section 4 we give the details of the robust parsing algorithm using a cost function defined in terms of null links. In Section 5 we explain how the pruning techniques of [5] can be extended to accommodate null links. Finally, in Section 6 we present the results of our experiments with the robust parser on the Switchboard corpus.

## 2  Link Grammar Concepts and Notation

In this section we briefly summarize the relevant concepts and notation of link grammar. The figure below represents one of the parses produced by a link grammar parser on the input sentence "Despite newspaper reports to the contrary, Mary handles herself with extreme confidence." The labelled arcs connecting words to other words on their left or right are called *links*. A valid parse is called a *linkage*.

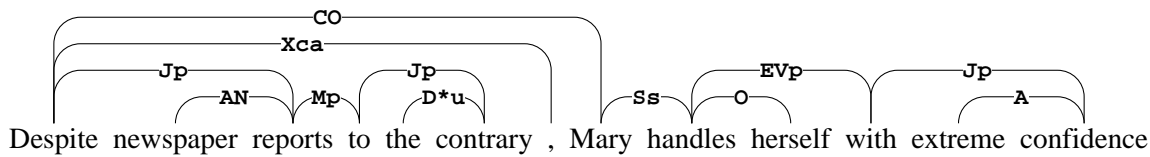

In this figure we see that `confidence` is linked on the left to the adjective `extreme` with a link labelled `A`, denoting an adjective. (We can also say that `extreme` is linked on the right to `confidence`.) Words can have multiple links; for example, The word `handles` has a singular noun (`Ss`) link to its left and object (`O`) and prepositional phrase (`EVp`) links to its right.

A link grammar is defined by a dictionary comprising a vocabulary and definitions of the vocabulary words. These definitions describe how the words can be used in linkages. A word's definition can be thought of as a list of *disjuncts*. Each disjunct $d$ is represented by two ordered lists, written as
$$d = (\,(l_1, l_2, \ldots, l_m)\ (r_n, r_{n-1}, \ldots, r_1)\,)$$
where $l_i$ are *left connectors* and $r_j$ are *right connectors*. (Connector names typically begin with one or more upper case letters followed by sequences of lower case letters and *'s.) For example, from the above linkage we can see that one of the disjuncts in the dictionary definition of `handles` must be ( (`Ss`) (`EVp,O`) ).

A word $W$ with disjunct $d = (\,(l_1, l_2, \ldots, l_m)\ (r_n, r_{n-1}, \ldots, r_1)\,)$ can be linked to other words by connecting each $l_i$ to right connectors of words on $W$'s left and also connecting each $r_j$ to left connectors of words on $W$'s right. Links are only permitted between *matching* connectors.

A linkage is specified by choosing a disjunct for each word in the sentence from the word's definition and linking every connector in the disjunct with a connector of a different word's disjunct. A linkage is valid if and only if it meets the following criteria:

Connectivity: The graph of links and words is connected.

Planarity: When drawn above the sentence, the links do not cross.

Ordering: For disjunct $d = (\,(l_1, l_2, \ldots, l_m)\ (r_n, r_{n-1}, \ldots, r_1)\,)$ of word $W$, $l_i$ and $r_j$ are linked to words whose distances from $W$ are monotonically increasing in $i$ and $j$.

Exclusion: No two links connect the same pair of words.

While the use of disjuncts simplifies the mathematical analysis and parsing algorithms for link grammars, it is cumbersome to express actual grammars in these terms. For this reason the linking requirements of a word are often expressed as a logical formula involving connectors and the operators & and or. As a simple example, the formula

(D- & (O- or S+))

represents the two disjuncts ( (D) (S) ) and ( (D,O) () ). The links attached to each word in a linkage must *satisfy* that word's formula. That is, the connectors must make the logical expression represented by the formula true, with the or operator understood to be exclusive.

A convenient data structure for storing and manipulating a link grammar dictionary represents a formula as an *expression tree*, with each node in the tree being either an or-node, an &-node, or a leaf representing a connector. To prepare for parsing a sentence, the expression tree for each word in the sentence is first pruned to eliminate connectors that cannot possibly participate in any linkage. The pruned expression tree is then expanded into a list of disjuncts before *power pruning* is carried out to eliminate disjuncts that necessarily violate one or more of the structural constraints that a valid linkage must obey. After pruning, the parsing algorithm itself is carried out. In Sections 4 and 5 we give the details on the parsing and pruning algorithms that are needed to explain the robust algorithm. In [5] the relationship between link grammar and other grammatical formalisms is discussed.

## 3 The Cost of a Linkage

The approach that we take to robust parsing uses the notion of *cost*. Informally, the idea is to define a set of generalized parse trees with respect to the grammar and the assignment of a number $cost(P \,|\, S) \geq 0$ to each generalized parse $P$ of a sentence $S$. The task of the robust parsing algorithm is to find all generalized parses having minimum cost. This, of course, is not a new idea; many approaches to robust parsing can be viewed in these terms. Some algorithms minimize the number of ungrammatical "islands," while statistical parsing algorithms for unrestricted text generally use the cost function $cost(P \,|\, S) = -\log \Pr(P \,|\, S)$ where $\Pr(P \,|\, S)$ is given by a probabilistic model. It is important to note that the objective of minimizing a cost function is only a heuristic. For any cost function there are likely to be examples for which "optimal" parses have non-minimal cost.

In this section we discuss two definitions of cost for link grammars. The first is called edit distance. The second, which introduces null links, is more restricted in the generalized parses it allows. We have been unable to devise a practical robust parsing algorithm for the edit distance, and thus we only mention it briefly as motivation. We have, however, been successful in designing an efficient algorithm using the more restricted definition of cost.

## 3.1 Edit distance

To define a cost function, we first need to define the set of generalized parses that are allowed. The definition of a linkage given in Section 2 requires that the links obey the planarity, connectivity, ordering, and exclusion properties. A set of links that obeys all of these conditions will be called *structurally sound*. If, in addition, the links satisfy the formula of each word in the sentence, the linkage will be called *legal*. We will assign a cost to any structurally sound linkage.

The *edit distance* of a structurally sound linkage is the minimum number of links and words that need to be added, deleted, or renamed in order to create a collection of one or more legal linkages. The edit distance of a string of words is defined to be the smallest edit distance of all structurally sound linkages connecting those words.

To illustrate these definitions, suppose that the following linkage is legal:

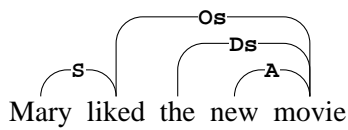

Then the edit distance of the string "liked the new movie" is no more than two, since the linkage above can be constructed from the following structurally sound linkage by adding a word and a link:

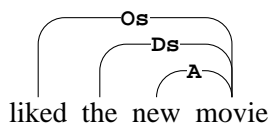

Similarly, the string "Mary liked the Fellini movie" has edit distance no greater than one if the word "Fellini" is not in the dictionary. The string "Mary liked the new movie John liked the new movie" also has edit distance no greater than one since the structurally sound linkage can be reduced to two legal linkages by removing a single link. In general, the edit distance of a string of $n$ words is no more than $2n - 1$. This is because the linkage obtained by linking only adjacent words can be disassembled into the empty linkage by deleting $n - 1$ links and all $n$ words.

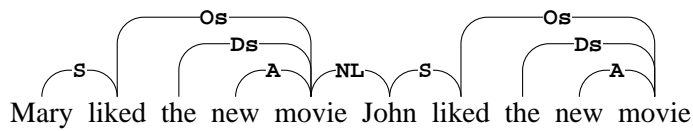

Using the original link grammar parsing methods, it is not too difficult to design an algorithm that takes $O(n^3)$ steps to calculate the edit distance of a string of $n$ words. However, we have been unable to find such an algorithm that will run efficiently for a significantly large grammar. The primary reason is that the pruning methods that allow the standard link grammar parsing algorithm to run efficiently do not have natural counterparts to prune the space of structurally sound linkages. However, by using a more restricted notion of edit distance we can design an efficient robust parsing algorithm.

## 3.2 Null links

A *null link* is an unlabeled link connecting adjacent words. The following linkage has two null links, drawn as dashed arcs.

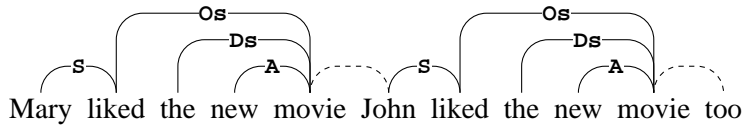

If both of these null links are deleted, then we obtain two legal linkages together with the disconnected word "too." If a structurally sound linkage has the property that removing its null links results in a collection of one or more legal linkages together with zero or more isolated words, we say that the linkage is *chained*. We define a cost to each chained linkage equal to the number of null links it contains. The cost of a string of words is defined to be the minimum cost of all chained linkages connecting those words. By this measure, the cost of a string of $n$ words can be no greater than $n-1$, and it is zero only if the string is grammatical. (We do not include a dependence on the number of isolated words since there is no *a priori* reason to favor one chained linkage over another, if they have the same number of null links, simply because it has fewer isolated words when disassembled.)

The set of chained linkages is the same as the set of legal linkages in an extended grammar. To construct this grammar we create a special connector name NL that does not appear in the original dictionary. For each disjunct

$$d = (\,(l_1, l_2, \ldots, l_m)\,(r_n, r_{n-1}, \ldots, r_1)\,)$$

in the original dictionary $\mathcal{D}$ we construct the following three disjuncts in the extended dictionary:

$$\begin{aligned} d' &= (\,(l_1, l_2, \ldots, l_m)\,(r_n, r_{n-1}, \ldots, r_1, \mathtt{NL})\,) \\ d'' &= (\,(\mathtt{NL}, l_1, l_2, \ldots, l_m)\,(r_n, r_{n-1}, \ldots, r_1)\,) \\ d''' &= (\,(\mathtt{NL}, l_1, l_2, \ldots, l_m)\,(r_n, r_{n-1}, \ldots, r_1, \mathtt{NL})\,). \end{aligned}$$

In addition, we add the following three disjuncts to the definitions of each word:

$$\begin{aligned} &(\,(\mathtt{NL})\,(\,)\,) \\ &(\,(\,)\,(\mathtt{NL})\,) \\ &(\,(\mathtt{NL})\,(\mathtt{NL})\,). \end{aligned}$$

This defines the extended dictionary $\mathcal{D}'$. Each link labeled NL in a linkage of the extended grammar corresponds to a null link. The removal of such links results in one or more legal linkages in the original grammar, together with zero or more disconnected words.

There are several advantages to using the restricted definition of cost described above. As we will show in the next two sections, there are natural extensions of the original parsing and pruning algorithms that accommodate null links. These extensions are easily implemented, and they lead to a robust parsing algorithm that is efficient and practical. Moreover, there is no need to explicitly construct the extended grammar just described; in fact, the grammar does not need to be modified in any way. Thus, null links represent virtual rather than physical links. Most importantly, as we indicate in Section 6, experimentation with the robust parser using null links has shown that it is capable of extracting relevant grammatical structure even in ungrammatical text such as transcriptions of spontaneous speech.

## 4   The Robust Parsing Algorithm

In this section we describe an algorithm for calculating the cost of a sentence. The algorithm determines the minimum number of null links necessary to parse the sentence by making three dynamic programming passes. In the first pass the input is parsed using the original algorithm which counts the number of legal linkages. If the sentence is grammatical then this is the only pass that is carried out. If not, a second pass is made to calculate the minimum number of null links necessary to parse each substring. Since the results of the first pass are memoized, only regions of the input sentence that are not grammatical need to be explored in this pass. The final pass counts the total number of parses with minimal cost.

The robust parsing algorithm is most easily understood by first considering the problem of counting *all* chained linkages, not just those with the fewest possible null links. To explain how this calculation is carried out we need to review the original link grammar parsing algorithm, which proceeds by recursively counting parses in a top-down fashion. Consider the situation after a link has been proposed between a connector $l'$ on word $L$ and a connector $r'$ on word $R$. For convenience, we define $l$ and $r$ to be $next[l']$ and $next[r']$ respectively.

In order to attach the words of the region $(L, \ldots, R)$ strictly between $L$ and $R$ to the rest of the sentence, there must be at least one link either from $L$ to some word in this region, or from $R$ to some word in this region (since no word in this region can link to a word outside of the $[L, \ldots, R]$ range, and something must connect these words to the rest of the sentence). Since the connector $l'$ has already been used in the solution being constructed, this solution must use the rest of the connectors of the disjunct in which $l'$ resides. The same holds for $r'$. The only connectors of these disjuncts that can be involved in the $(L, \ldots, R)$ region are those in the lists beginning with $l$ and $r$. (The use of any other connector on these disjuncts in this region would violate the ordering requirement.) In fact, all of the connectors of these lists must be used in this region in order to have a satisfactory solution.

Suppose that $l$ is not NIL. We know that this connector must link to some disjunct on some word in the region $(L, \ldots, R)$. The algorithm tries all possible words and disjuncts. Suppose it finds a word $W$ and a disjunct $d$ on $W$ such that the connector $l$ matches $left[d]$. We can now add this link to our partial solution. The situation is shown in the following diagram.

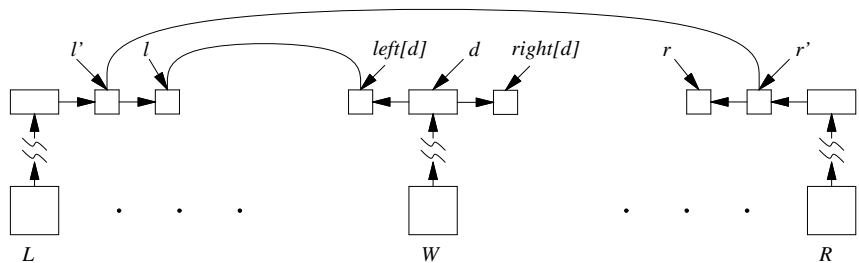

Here the square box above $L$ represents the data structure node corresponding to the word $L$. This points to a list of disjuncts, each of which is shown as a rectangular box above the word. The disjunct in turn points to two linked lists of connectors, shown here as small square boxes.

The algorithm determines if this partial solution can be extended to a full solution by solving two problems similar to the original problem. In particular, the word range $(L, \ldots, W)$ is extended using the connector lists beginning with $next[l]$ and $next[left[d]]$. In addition, the word range $(W, \ldots, R)$ is extended using the connector lists beginning with $right[d]$ and $r$. The case where $l = \text{NIL}$ is similar.

To explain the robust algorithm it will be helpful to repeat a fragment of the pseudocode representing the original algorithm as given in [5]. The function COUNT takes as input indices of two words $L$ and $R$, where the words are numbered from 0 to $N-1$, and pair of connector lists $l$ and $r$. COUNT returns the number of different ways to draw links among the connectors on the words strictly between $L$ and $R$, and among the connectors in lists $l$ and $r$.

PARSE
1  $t \leftarrow 0$
2  **for** each disjunct $d$ of word 0
3       **do if** $left[d] = \text{NIL}$
4            **then** $t \leftarrow t + \text{COUNT}(0, N, right[d], \text{NIL})$
5  **return** $t$

COUNT$(L, R, l, r)$
1  **if** $R = L + 1$
2     **then if** $l = \text{NIL}$ and $r = \text{NIL}$
3             **then return** 1
4             **else return** 0
5  **else** $total \leftarrow 0$
6     **for** $W \leftarrow L + 1$ **to** $R - 1$
7         **do for** each disjunct $d$ of word $W$
8               $\ldots$

The algorithm counts all linkages by recursively counting smaller and smaller regions. Lines 1–4 of the COUNT procedure handle the boundary conditions. If word $R$ is adjacent to $L$ then there is no solution unless $l = r = \text{NIL}$. This ensures that all of the connectors are used for every disjunct that participates in a linkage. If $L$ and $R$ are not neighboring words then the algorithm examines each intervening word $W$, beginning at line 6.

To explain how null links are introduced we need to first closely examine the situation where $l = \text{NIL}$ and $r \neq \text{NIL}$. In this case, to extend the linkage into the region $(L, \ldots, R)$, a word $W$ must be chosen to link with $r$. This leads to a test of whether the region $(L, \ldots, W)$ can be extended using the connector lists NIL for $L$ and $left[d]$ for some disjunct $d$ of $W$. Assuming a leftmost derivation, this process continues until either $left[d] = \text{NIL}$ or $W = L + 1$. If $left[d] = \text{NIL}$ then there is no solution in the original algorithm except when the boundary condition $W = L + 1$ holds. In the robust algorithm, however, there is always a way of continuing.

Note that in case $l = \text{NIL}$ when parsing with respect to the original dictionary $\mathcal{D}$, we can always replace $l$ by NL for the extended grammar $\mathcal{D}'$. Since NL connectors can only link adjacent words, this means that $L$ could only be connected to $W = L + 1$ with a disjunct $d$ whose connector $left[d]$ is NL in the extended grammar. The robust algorithm makes such links "virtually," without actually parsing the extended grammar. If $R > L + 1$ and $l = r = \text{NIL}$, the algorithm considers each disjunct $d$ of $W = L + 1$ having the property that $left[d] = \text{NIL}$. For each such disjunct, a null link is made between $L$ and $W$ and then the region $(W, \ldots, R)$ is parsed using the connector list beginning with *right*[*d*] for *W*. Another way of proceeding is to make a null link between *L* and *W* and parse the region (*W*,...,*R*) using the connector NIL for *W*. This corresponds to using the disjunct ((NL) (NL)) for *W* in the extended grammar.

The counting algorithm just outlined is given more formally in the pseudocode below. The code after line 12 is identical to that of the original algorithm.

PARSE
1  $t \leftarrow 0$
2  **for** each disjunct *d* of word 0
3     **do if** *left*[*d*] = NIL
4        **then** $t \leftarrow t + \text{COUNT}(0, N, right[d], \text{NIL})$
5  $t \leftarrow t + \text{COUNT}(0, N, \text{NIL}, \text{NIL})$
6  **return** *t*

COUNT(*L*, *R*, *l*, *r*)
 1  **if** $R = L + 1$
 2     **then if** *l* = NIL and *r* = NIL
 3        **then return** 1
 4        **else return** 0
 5  **elseif** *l* = NIL and *r* = NIL
 6     **then** *total* $\leftarrow 0$
 7        **for** each disjunct *d* of word $L + 1$
 8           **do if** *left*[*d*] = NIL
 9              **then** *total* $\leftarrow$ *total* $+ \text{COUNT}(L + 1, R, right[d], \text{NIL})$
10           *total* $\leftarrow$ *total* $+ \text{COUNT}(L + 1, R, \text{NIL}, \text{NIL})$
11        **return** *total*
12  **else** ...

Note that the boundary conditions in lines 1–4 are identical to those of the original algorithm. In the case where $l = r =$ NIL and $R > L + 1$ the original counting procedure returns 0. This is where null links are introduced in the robust algorithm.

This algorithm returns a number that is, in general, less than the number of legal linkages in the extended grammar; that is, it only counts a subset of all chained linkages. The reason is that the extended grammar introduces NL links symmetrically, and this symmetry is not "broken" by the parsing algorithm. For example, if one of the following linkages is legal in the extended grammar then all three of them are:

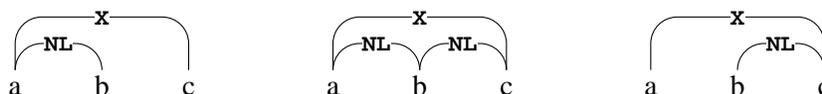

In the algorithm given above, however, a null link can be introduced in this case only as follows:

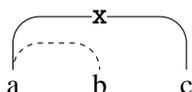

This inconsistency is easily remedied. The algorithms will return the same number of chained linkages if the original parsing algorithm for the extended grammar is modified so that in the case where $l = $ NL and $r = $ NL, a connection is only allowed to $l$.

Knowing the total number of chained linkages is of limited use. This collection of parses includes many that have little or no structural information, such as the linkage that joins each word with its immediate neighbors by null links. Using the ideas just described, however, we can count the number of linkages that have the minimal number of null links. We first modify the algorithm to compute the cost of each span. To do this, the increment operations are replaced by "min" operations. The following pseudocode demonstrates how this is carried out.

CALCCOST
1  $c \leftarrow \infty$
2  **for** each disjunct $d$ of word 0
3      **do if** $left[d] = $ NIL
4          **then** $c \leftarrow $ MIN$(c, $ COST$(0, N, right[d], $ NIL$))$
5  $c \leftarrow $ MIN$(c, $ COST$(0, N, $ NIL, NIL$))$
6  **return** $t$

COST$(L, R, l, r)$
1  **if** $R = L + 1$
2    **then if** $l = $ NIL and $r = $ NIL
3        **then return** 0
4        **else return** $\infty$
5  **elseif** $l = $ NIL and $r = $ NIL
6    **then** $cost \leftarrow \infty$
7        **for** each disjunct $d$ of word $L + 1$
8          **do if** $left[d] = $ NIL
9            **then** $cost \leftarrow $ MIN$(cost, $ COST$(L + 1, R, right[d], $ NIL$) + 1)$
10        $cost \leftarrow $ MIN$(cost, $ COST$(L + 1, R, $ NIL, NIL$) + 1)$
11        **return** $cost$
12  **else** . . .

Note that while the cost is initialized to be infinite on line 1 of CALCCOST and line 6 of COST, the cost of a string of $n$ words can be no greater than $n - 1$. The code after line 12 is modified from the original code in the obvious way. If a region $(L, \ldots, R)$ with connector lists $l$ and $r$ is grammatical, this fact will have been recorded (memoized) during the first pass which carries out the standard algorithm. In this case the cost is zero, and there is no need to carry out the second pass on the span. Thus, the second pass only needs to explore those regions that are ungrammatical.

In the third pass, the number of chained linkages with minimal cost is calculated in a manner similar to the way in which the second pass is carried out. However, rather using than a "min" operation, a count is incremented if a given span has minimum cost. The details are omitted.

## 5 Pruning Techniques

The original link grammar parsing algorithm is made practical by means of several pruning techniques, including expression pruning, the fast-match data structure, power pruning, and conjunction pruning. This section shows how each of these techniques has a simple and natural extension to the robust parsing algorithm.

### 5.1 Pruning

Suppose that a word $W$ has a disjunct $d$ with a connector C in its list of right connectors. In order for $d$ to be used for $W$ in a linkage of a given sentence, some word to the right of $W$ must have a disjunct with a connector that matches C in its left list. This simple observation is the basis for an algorithm to prune the set of disjuncts that are considered by the parsing algorithm.

The pruning algorithm alternately makes sequential left-to-right and right-to-left passes through the words in the sentence. In a left-to-right pass, a set of right connectors is maintained, and is initially empty. The pruning step for word $W$ consists of examining each disjunct $d$ of $W$ that has survived previous pruning passes. If a left connector of $d$ does not match any connector in the set, then $d$ is discarded. Otherwise, the connectors in the right list of $d$ are added to the set. After all disjuncts of $W$ have been processed in this way, the algorithm advances to the next word. The right-to-left pass is analogous.

Pruning can be carried out more efficiently when applied to the entire expression tree representing the formula of a word. In a left-to-right expression pruning pass, each left connector in a word's expression tree is checked to see if it has a match in the set of right connectors. If not, the connector and those nodes that depend on it through & relations are pruned from the tree.

When parsing with null links, the pruning algorithm can be used without modification. Viewing null links in terms of the extended grammar, it can be seen that a disjunct will never be pruned because of the lack of a matching NL connector (excepting boundary cases where there is no word to the left or right). In other words, applying pruning to the original grammar and then extending the surviving disjuncts with NL connectors results in the same set of disjuncts as applying the pruning algorithm to the extended grammar.

### 5.2 The fast-match data structure

The fast-match data structure is used in the parsing algorithm to quickly determine which disjuncts of a word might possibly match a given left or right connector. The data structure uses two hash tables for each word in the sentence. The disjuncts that survive pruning are hashed, and disjuncts that hash to the same location are maintained in a linked list. The left table for a word uses a hash function that depends only on the first connector on the left list of the disjunct. The right table has an analogous hash function. When parsing a span $(L, \ldots, R)$ by attempting to link the connector lists $l$ and $r$ to an intermediate word $W$, the fast-match data structure allows a list of candidate disjuncts for $W$ to be quickly obtained by forming the union of the lists obtained from the left table of $W$ by hashing $l$ and the right table of $W$ by hashing $r$.

A simple modification to the fast-match data structure can be made to accommodate null links. As discussed in the previous section, when null links are constructed "virtually" rather than by explicitly parsing the extended grammar, a null link is only made in the case where the region $(L, \ldots, R)$ is extended using connector lists $l$ and $r$ which are both NIL. In this case, the only disjuncts of word $W = L + 1$ that are considered are those with empty left connector lists. Thus, to modify the fast-match data structure it is only necessary to maintain, for each word, a list of disjuncts $d$ for which $left[d] = $ NIL.

## 5.3 Power pruning

After expression pruning is carried out, each word's pruned expression tree is expanded into a list of disjuncts. At this point power pruning is carried out in an attempt to enforce the structural constraints imposed on valid linkages, including the ordering, connectivity, planarity, and exclusion properties. As with pruning, power pruning is carried out by making alternating left-to-right and right-to-left passes through the sentence; the details are given in [5].

The power pruning algorithm eliminates some disjuncts by observing that in order for a linkage to be connected, it is necessary that any two connectors between *non-neighboring* words cannot both be the last connectors of their respective lists. This condition is relaxed when parsing with null links. In fact, a null link is *only* formed after two non-neighboring words are connected using the last connectors in their lists. All of the other conditions that power pruning checks remain valid. Thus, it is a simple matter to modify the power pruning algorithm to allow for the possibility of null links.

In the case where the sentence contains a conjunction, a variation of the power pruning algorithm enables pruning of disjuncts before *fat connectors* are built [5]. This variation uses the notion of a *deletable region*. A substring of words is said to be a deletable region if its removal would result in a grammatical sentence. For example, in the sentence *the dog and cat ran*, both *dog and* and *and cat* are deletable regions. In essence, conjunction pruning proceeds by parsing the sentence but allowing for the excision of one or more deletable regions. Each disjunct that participates in such a parse is marked; all unmarked disjuncts can be pruned away. This strategy is again easily extended to the situation where we allow null links. This is done by simply using the robust parsing algorithm when marking disjuncts, and allowing any region of words to be deletable.

## 6 Experimental Results

The Switchboard corpus was created to allow standard evaluations and provide a stable research environment for large vocabulary continuous speech recognition. The corpus contains an abundance of phenomena associated with conversational language: non-standard words, false starts, stutters, interruptions by other speakers, partial words, grammatical errors, changes in grammatical structure mid-sentence, hesitation sounds, and non-speech sounds.

This section describes the results of an experiment in which we applied our robust parser to a randomly chosen subset of the Switchboard corpus. Our purpose in doing this experiment was twofold. First, we wanted to determine whether the robust parsing algorithm, built upon a grammar that attempts to describe "correct" English, would be able to glean useful structure from such "incorrect" text. Secondly, we wanted to measure the time efficiency of the parser. It was not our purpose to give an objective measure of the quality of the grammar, or to make a direct comparison with other robust parsing methods.

| sentences | # not skipped | 909 (56%) |
|---|---|---|
| | # skipped (too short) | 706 (44%) |
| | # grammatical | 202 (22%) |
| | # with unknown words | 115 (13%) |
| words | # not skipped | 12047 (89%) |
| | # skipped (too short) | 1452 (11%) |
| linkages | Average # per sentence | 114 |
| | Average # of null links | 1.99 |
| time | Total wall time (seconds) | 3143 |
| | Maximum for one sentence | 273 |

Other than the entries listing the number of sentences and words skipped, skipped sentences do not participate in the calculation of the other entries. The total wall time is the elapsed time needed to compute and sort all linkages for each sentence on a DEC AXP 3000/600 Alpha workstation.

Table 1: The results of the robust parser on Switchboard data

Before parsing the text, we removed all punctuation and case information. We also combined some commonly occuring word pairs (such as "you know") into single words; however, the dictionary was not modified to account for these new words. These changes make the problem of parsing the Switchboard corpus closer to that of analyzing speech.

One problem that emerged in trying to apply the parser to this corpus was the lack of sentence boundaries. The parser expects its input to be broken into "sentence size" blocks. A natural approach is to use changes of speaker identity to partition the input. We will call such a block of text an *utterance*. Some utterances are very long, and require a lot of time and memory to parse if fed to the parser as a single sentence. We explored two ways to deal with this problem. The first (reported below) is merely to split long utterances into shorter groups of words (that we will call sentences) that can be parsed using reasonable resources. This loses some grammatical information. But with even an unsophisticated splitting algorithm the parser still seems to discover useful structure. Another method to alleviate the high cost of parsing long sentences is to limit the length of the longest link permitted.

In the experiment reported here, we chose 1500 utterances at random from the Switchboard corpus. Each utterance was split into pieces no longer than 25 tokens each. (The number of words in an sentence after the splitting can still be greater than 25 because the parser considers some tokens, such as possessives, to be more than one word.) We also skipped all sentences with fewer than four words. The results of applying the robust parser to this data are summarized in Table 1.

Somewhat surprisingly, given that only 22% of the corpus's sentences were found to be grammatical, only two null links were needed on average to parse the corpus. In order to better understand this statistic we randomly permuted the 13,499 words in the 1617 sentences we previously used, preserving the lengths of the sentences. Of these randomly generated sentences, only 8 (0.89%) were grammatical, and on average, 6 null links were needed to parse them.

Perhaps the most conspicuous feature that distinguishes the utterances in the Switchboard data from more grammatical text is the frequent occurrence of repeated words. The following example contains two repeats of the word WE'VE, in addition to a false start. (A description of the connectors used is given in an appendix.) The robust parser successfully ignores the false start by enforcing the constraint of subject-verb agreement. The three occurrences of WE'VE are interchangeable, and this results in three equivalent parses for every use of this word.

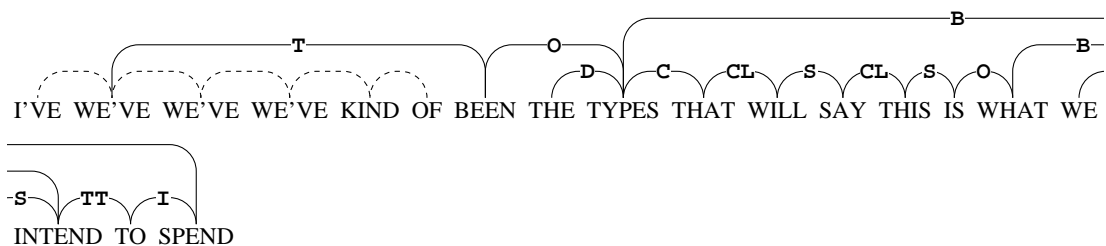

It is also common in this data for an utterance to include an embedded phrase that is grammatical in isolation but not with respect to the larger utterance. As the following example demonstrates, the robust parser can handle this by connecting the embedded phrase to the larger utterance with a null link. There were 5 other linkages of this utterance that are essentially equivalent to the one shown below.

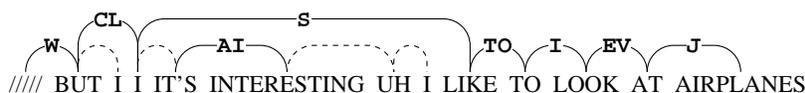

In many cases the utterance is ungrammatical, yet by minimizing the number of null links the parser is able to recover the relevant structure. The following example fails to be grammatical because of false starts and the sequence of words INSTEAD OF THE PROSECUTING.

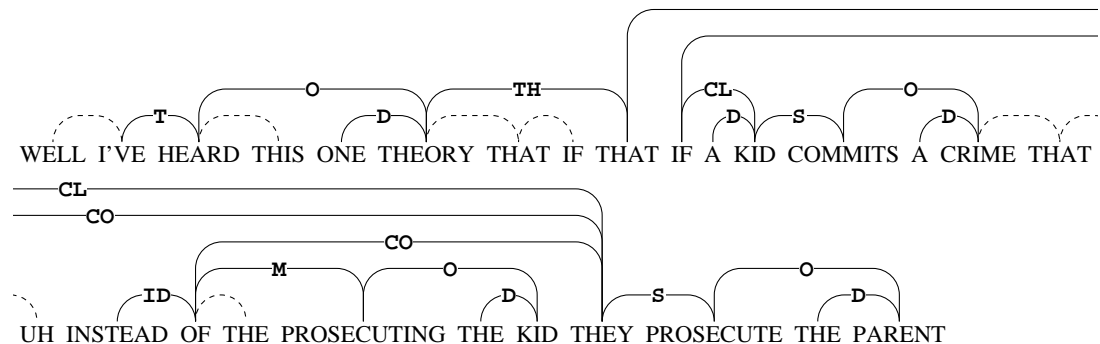

### 6.1 Random sentences

We now present parses of ten randomly chosen sentences from the corpus.

1. UH I HAVEN'T BEEN TO ANY TORONTO GAMES YET

There are 6 parses for this sentence having a single null link.

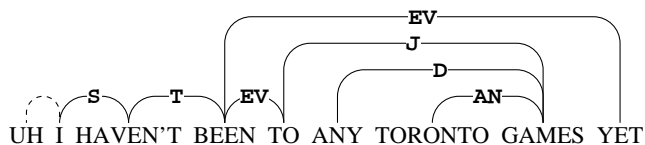

The word TORONTO is not in the dictionary. When a word is not defined, the parser assigns a default list of disjuncts to the word. In this case the context is used to correctly determine that the unknown word functions as an adjective.

2. WELL IT'S IT IT YEAH IT'S A LITTLE BIT LIKE ANY OTHER SPORT YOU_KNOW

This sentence has six linkages with four null links.

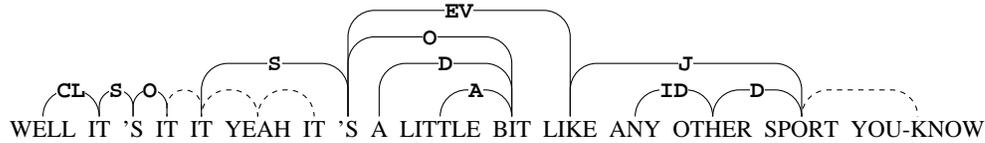

3. UH ARE YOU PRESENTLY LOOKING UH FOR A USED CAR

One linkage with two null links.

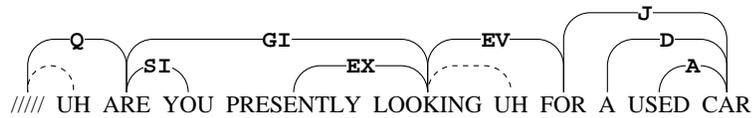

4. WHEN I WAS IN I USED TO LIVE IN CALIFORNIA

One linkage with no null links.

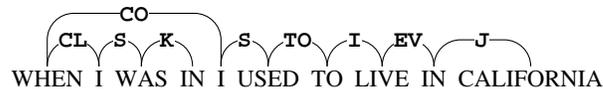

5. IS IS THE KMART THERE

Two linkages differing only in which instance of IS is used.

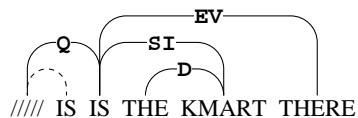

The word KMART is not in the dictionary, but the unknown word mechanism takes advantage of the context to choose an appropriate disjunct.

6. WELL I THINK UH ONE THING THAT WE'LL SEE IS THAT WE WON'T BE EDUCATING EVERYBODY

One linkage with two null links.

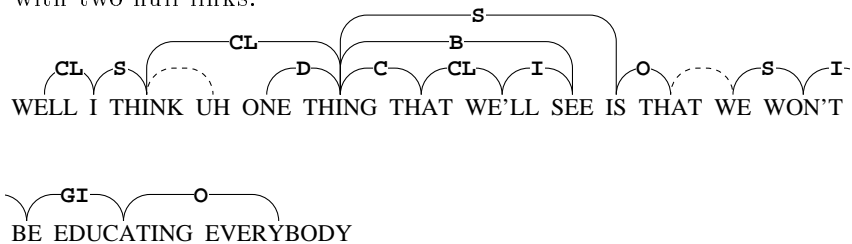

The parser incorrectly splits this sentence into WELL I THINK ONE THING THAT WE'LL SEE IS THAT and WE WON'T BE EDUCATING EVERYBODY.

7. AND THEY DON'T KNOW A LOT ABOUT THAT <NOISE>

Five linkages with one null link. They differ in whether ABOUT modifies A LOT or KNOW, whether A LOT is an object or an adverb and whether or not to treat A LOT as an idiom. The correct linkage is shown below.

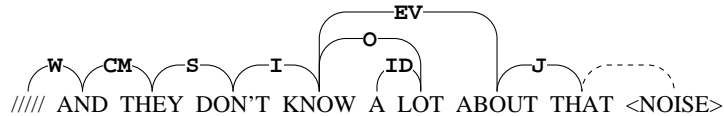

8. I_MEAN BUT IT WOULD TO BE AN EMERGENCY TYPE THING

One linkage with two null links.

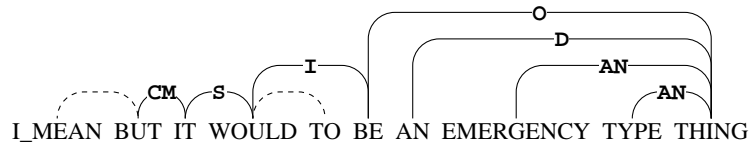

9. RIGHT KIND OF LIKE UH UH BAREFOOT CRUISES TYPE OF THING YEAH UH_HUH

Five similar linkages with nine null links.

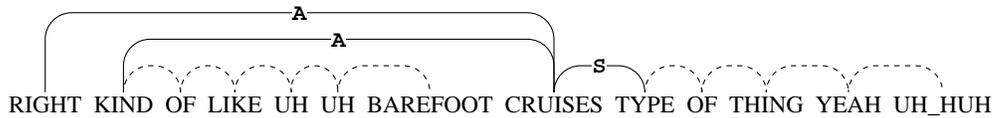

10. I DON'T HAVE ANY EITHER AND I'VE BEEN A MARRIED FOR ELEVEN YEARS SO

Two linkages with two null links. The following linkage treats AND as a conjunction.

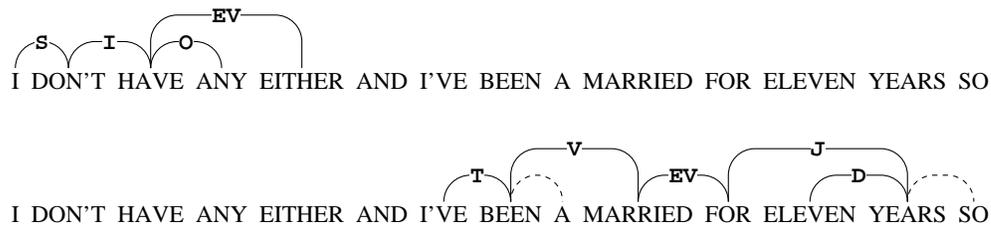

## 6.2 Analysis

The mixed-case version of the parser treats words beginning with uppercase characters as proper nouns. However, since there is no case information available in our data most proper nouns must be treated as unknown words. The use of a generic definition for unknown words often interprets these words correctly, taking clues from the surrounding context.

Minimizing the number of null links in a linkage, while a fairly good heuristic to produce good parses, is not foolproof. For example, we might interpret example 4 as a false start of WHEN I WAS in followed by I USED TO LIVE IN CALIFORNIA. The parser cannot mimic this interpretation because to do so would require more than the minimum number of null links. Example 6 demonstrates a deficiency of the grammar, and suggests that a learning algorithm would be useful to infer new word usages and grammatical relations. Similarly, example 8 shows that the parser can not infer missing words. We would like the parser to realize that HAVE should be inserted between WOULD and TO. As we see from example 9, the parser can utterly fail at producing useful structure.

The robust parser has limitations on the input it can parse. For the Switchboard domain, splitting long sentences into shorter ones does not seem to cause significant problems but on other domains it might. Limiting the length of the longest link permitted can allow the system to parse long utterances in reasonable time, with realistic memory limitations. While the resulting linkages might not be as good as ones produced without any limitation, the tradeoff of time and space versus accuracy may be worthwhile for many applications.

# References

[1] J. Godfrey, E. Holliman, and J. McDaniel. Switchboard: Telephone speech corpus for research development. In *Proc. ICASSP-92*, pages I–517–520, 1992.

[2] A. Lavie and M. Tomita. Parsing unrestricted sentences using a generalized LR parser. In *Proceedings of the Third International Workshop on Parsing Technologies*, pages 123–134, 1993.

[3] K. J. Lee, C. J. Kweon, J. Seo, and G. C. Kim. A robust parser based on syntactic information. In *Proceedings of the 7th Conference of the European Chapter of the Association for Computational Linguistics*, 1995.

[4] G. Lyon. Syntax-directed least-errors analysis for context-free languages: A practical approach. *Communications of the ACM*, 17(1):3–14, Jan 1974.

[5] D. D. K. Sleator and D. Temperley. Parsing English with a link grammar. Technical Report CMU-CS-91-196, School of Computer Science, Carnegie Mellon University, 5000 Forbes Avenue, Pittsburgh, PA 15213, 1991.

## A  Descriptions of Connectors

- **A**  Connects prenominal adjectives to nouns.
- **AN**  Connects noun-modifiers to nouns.
- **B**  Used in a number of situations, involving relative clauses and questions.
- **C**  Connects nouns to relative clauses.
- **CL**  Connects conjunctions and certain verbs with subjects of clauses.
- **CM**  Connects conjunctions and certain verbs with subjects of clauses.
- **CO**  Used to connect "openers" to subjects of clauses.
- **D**  Connects determiners to nouns.
- **DS**  Connects determiners to number expressions.
- **EV**  Connects verbs and adjectives to modifying phrases like adverbs, prepositional phrases or "words," certain conjunctions, and other items.
- **EX**  Used for adverbs modifying verbs which precede the verb.
- **GI**  Connects verbs that take present participles with present participles.
- **I**  Connects certain verbs with infinitives.
- **ID**  Idioms.
- **J**  Connects prepositions to their objects.
- **K**  Connects certain verbs with participles like "in," "out," "up," and the like.
- **M**  Connects nouns to various kinds of post-nominal modifiers, such as prepositional phrases, collapsed relatives and possessive relatives.
- **O**  Connects transitive verbs to direct or indirect objects.
- **Q**  Used in questions.
- **R**  Connects certain verbs and adjectives to question-words, forming indirect questions.
- **S**  Connects subject-nouns to finite verbs.
- **SI**  Used in subject–verb inversion.
- **SXI**  Connects non-referential subjects "it" and "there" to invertible verbs in questions with subject-verb inversion.
- **T**  Connects forms of "have" with past participles.
- **TH**  Connects words that take "that clause" complements with the word "that."
- **TT**  Connects certain verbs to infinitival complements, when the verb is also taking a direct object. the word "that."
- **TO**  Connects verbs and adjectives which take infinitival complements to the word "to."
- **Q**  Connects auxiliaries in simple subject–verb inversion questions when there is no preceding object.
- **V**  Used to connect passive participles to forms of "be."

17